\pdfoutput=1
\PassOptionsToPackage{dvipsnames}{xcolor} 
\documentclass[11pt]{article}

\usepackage[]{eacl2023}

\usepackage{times}
\usepackage{latexsym}
\usepackage{inconsolata}
\usepackage{tcolorbox}
\usepackage{booktabs}
\usepackage{multirow}
\usepackage{arydshln}
\usepackage{float}
\usepackage{tabularx}
\usepackage{caption,subcaption}
\usepackage{colortbl}
\usepackage{amssymb}%
\usepackage{pifont}%
\newcommand{\cmark}{\ding{51}}%
\newcommand{\xmark}{\ding{55}}%

\usepackage[T1]{fontenc}

\usepackage[utf8]{inputenc}

\definecolor{Gray}{gray}{0.92}
\definecolor{LightGray}{gray}{0.96}
\definecolor{LightCyan}{rgb}{0.92,0.968,0.968}
\definecolor{amber}{rgb}{1.0, 0.75, 0.0}%
\definecolor{neuesrot}{RGB}{207, 103, 102}%
\definecolor{lightblue}{RGB}{100, 181, 246}%
\definecolor{lightgreen}{RGB}{129, 199, 132}%
\newcolumntype{Y}{>{\centering\arraybackslash}X}

\usepackage{microtype}

\newcommand{\rparagraph}[1]{\vspace{1.6mm}\noindent\textbf{#1.}}
\newcommand{\sparagraph}[1]{\vspace{0.0mm}\noindent\textbf{#1.}}

\usepackage{xcolor}
\usepackage{tikz} 
\usepackage{pgfplots}
\usepackage{gb4e}
\noautomath
\usepackage[inline]{enumitem}

\newtheorem{finding}{Finding}

\usepackage{graphicx}
\usepackage{adjustbox}

\title{Can Pretrained Language Models (Yet) Reason Deductively?}

\author{
Zhangdie Yuan$^{\diamondsuit}\Thanks{~Indicates equal contribution.}$ ,
Songbo Hu$^{\spadesuit *}$, 
Ivan Vuli\'{c}$^\spadesuit$, 
Anna Korhonen$^\spadesuit$, 
Zaiqiao Meng$^\clubsuit$$^\spadesuit$\Thanks{~Corresponding author.} \\
$^\diamondsuit$Department of Computer Science and Technology, University of Cambridge \\
$^\spadesuit$Language Technology Lab, University of Cambridge \\
$^\clubsuit$School of Computing Science, University of Glasgow\\
  \texttt{$^{\diamondsuit \spadesuit}$\{zy317,sh2091,iv250,alk23\}@cam.ac.uk} \\
  \texttt{$^\clubsuit$zaiqiao.meng@glasgow.ac.uk}
  }
 
\pgfplotsset{compat=1.17}
\begin{document}
\maketitle
\begin{abstract}

Acquiring factual knowledge with Pretrained Language Models (PLMs) has attracted increasing attention, showing promising performance in many knowledge-intensive tasks. Their good performance has led the community to believe that the models do possess a modicum of reasoning competence rather than merely memorising the knowledge. In this paper, we conduct a comprehensive evaluation of the learnable \textit{deductive} (also known as explicit) \textit{reasoning capability} of PLMs. Through a series of controlled experiments, we posit two main findings. 
\begin{enumerate*}[label=(\roman*)]
\item PLMs inadequately generalise learned logic rules and perform inconsistently against simple adversarial surface form edits.
\item While the deductive reasoning fine-tuning of PLMs does improve their performance on reasoning over unseen knowledge facts, it results in catastrophically forgetting the previously learnt knowledge. 
\end{enumerate*}
Our main results suggest that PLMs cannot yet perform reliable deductive reasoning, demonstrating the importance of controlled examinations and probing of PLMs' deductive reasoning abilities; we reach beyond (misleading) task performance, revealing that PLMs are still far from robust reasoning capabilities, even for simple deductive tasks.

\end{abstract}

\section{Introduction}

Pretrained Language Models (PLMs) such as BERT~\cite{devlin-etal-2019-bert} and RoBERTa~\cite{liu2019roberta} have orchestrated tremendous progress in NLP across a large variety of downstream applications. For knowledge-intensive tasks in particular, these large-scale PLMs are surprisingly good at memorising factual knowledge presented in pretraining corpora~\cite{petroni2019language,jiang2020can} and infusing knowledge from external sources~\cite[\textit{among others}]{wang2021k,Zhou:2022acl}, demonstrating their effectiveness in learning and capturing knowledge.

\begin{figure}[!t]
    \centering
    \includegraphics[width=0.98\linewidth]{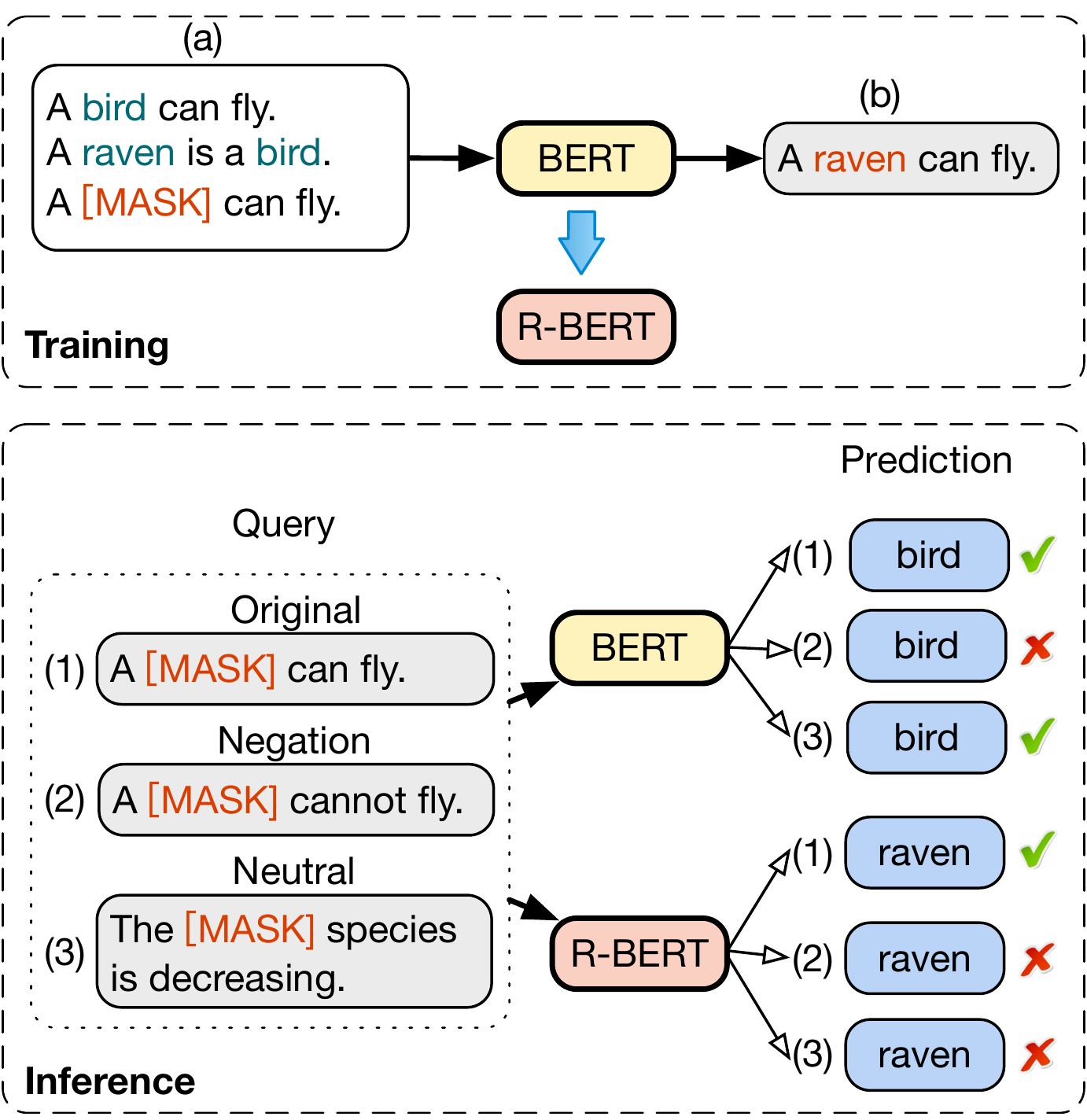}    \caption{Training and inference for deductive reasoning. Given the explicit premises (a), the input BERT model is trained to get transformed into a reasoner R-BERT model by deductively predicting a previously unseen conclusion (b). This inference process requires R-BERT to understand factual knowledge and interpret rules (e.g. taxonomic relations), intervening directly in the deduction process.}
    \label{fig:fig1}
\end{figure}

Automatic reasoning, a systematic process of deriving previously unknown conclusions from given formal representations of knowledge~\cite{mccarthy1959programs, newell1956logic}, has been a long-standing goal of AI research. In the NLP community, a modern view of this problem~\cite{clark2020transformers}, where the formal representations of knowledge are substituted by the natural language statements, has recently received increasing attention,\footnote{Following~\citet{clark2020transformers}, we also define natural language rules as linguistic expressions of conjunctive implications, $ { condition } [\wedge { condition }]^{*} \rightarrow { conclusion }$, with the semantics of logic programs with negations~\cite{apt1988towards}.} yielding multiple exploratory research directions: mathematical reasoning~\cite{rabe2020mathematical}, symbolic reasoning~\cite{yang2021learning}, and commonsense reasoning~\cite{li2019teaching}. Impressive signs of progress have been reported in teaching PLMs to gain reasoning ability rather than just memorising knowledge facts~\cite{kassner2020pretrained, talmor2020leap}, suggesting that PLMs could serve as effective reasoners for identifying analogies and inferring facts not explicitly/directly seen in the data~\cite{kassner2020pretrained, ushio2021bert}.

In particular, \emph{deductive reasoning}\footnote{This type of reasoning is also often referred to as \textit{explicit reasoning} in the literature \cite{broome2013rationality,Aditya:2018aaai}.} is one of the most promising directions~\cite{sanyal2022fairr, talmor2020leap,li2019teaching}. By definition, deduction yields valid conclusions, which must be true given that their premises are true~\cite{johnson1999deductive}.
In the NLP community, given all the premises in natural language statements, some large-scale PLMs have shown to be able to deductively draw appropriate conclusions under proper training schemes ~\cite{clark2020transformers,talmor2020leap}.
Figure~\ref{fig:fig1} shows an example of the training and inference processes of deductive reasoning.

Despite promising applications of PLMs, some recent studies have pointed out that they could only perform a shallow level of reasoning on textual data~\cite{helwe2021reasoning}. Indeed, PLMs can be easily affected by mispriming~\cite{misra2020exploring} and still hardly differentiate between positive and negative statements (i.e., the so-called \textit{negation issue})~\cite{ettinger2020bert}. However, given that some evidence suggests that PLMs can learn factual knowledge beyond mere rote memorisation~\cite{heinzerling2020language} and their limitations~\cite{helwe2021reasoning}, it is natural to ask, ``\emph{Can the current PLMs potentially serve as reliable deductive reasoners over factual knowledge?}'' To answer it, as the main contribution of this work, we conduct a comprehensive experimental study on testing the learnable deductive reasoning capability of the PLMs.

In particular, we test various reasoning training approaches on two knowledge reasoning datasets. Our experimental results indicate that such deductive reasoning training of the PLMs (e.g., BERT and RoBERTa) yields strong results on the standard benchmarks, but it inadequately generalises learned logic rules to unseen cases. That is, they perform inconsistently against simple surface form perturbations (e.g., simple synonym substitution, paraphrasing or negation insertion), advocating a careful rethinking of the details behind the seemingly flawless empirical performance of deductive reasoning using the PLMs. We hope our work will inspire further research on probing and improving the deductive reasoning capabilities of the PLMs. Our code and data are available online at \url{https://github.com/cambridgeltl/deductive_reasoning_probing}.

\section{Related Work}

\sparagraph{Knowledge Probing, Infusing, and Editing with PLMs}
PLMs appear to memorise (world) knowledge facts during pretraining, and such captured knowledge is useful for knowledge-intensive tasks~\cite{petroni2019language,petroni2021kilt}. A body of recent research has aimed to
\begin{enumerate*}[label=(\roman*)]
\item understand how much knowledge PLMs store, i.e., \textit{knowledge probing}~\cite{petroni2019language,meng2021rewire};
\item how to inject external knowledge into them, i.e., \textit{knowledge infusing}~\cite{wang2019kepler,meng2021mixture}; and 
\item how to edit the stored knowledge, i.e. \textit{knowledge editing}~\cite{de2021editing}.
\end{enumerate*}
In particular, \citet{de2021editing} have shown that it is possible to modify a single knowledge fact without affecting all the other stored knowledge. However, some empirical evidence suggests that existing PLMs generalise poorly to unseen sentences and are easily misled~\cite{kassner2020negated}.\footnote{For instance, if we add the \textit{talk} token into the statement ``Birds can [MASK].'' (i.e. ``Talk. Birds can [MASK].''), the PLM might be misled by the added token and predict \textit{talk} rather than the originally predicted \textit{fly} token~\cite{kassner2020negated}.} Moreover, this body of research focuses only on investigating how to \textit{recall or expose} the factual and commonsense knowledge that has been encoded in the PLMs, rather than exploring their capabilities of deriving previously unknown knowledge via \textit{deductive reasoning}, as done in this work.

\rparagraph{Knowledge Reasoning with PLMs} 
In recent years, PLMs have also achieved impressive progress in \emph{knowledge reasoning}~\cite{helwe2021reasoning}. For example, PLMs can infer a conclusion from a set of knowledge statements and rules~\cite{talmor2020leap,clark2020transformers}, with both the knowledge and the rules being mentioned explicitly and linguistically in the model input. Some generative PLMs, such as T5 \cite{Raffel:2019t5}, are even able to generate natural language proofs that support implications over logical rules expressed in natural language ~\cite{tafjord2021proofwriter}. In particular, some large PLMs, such as LaMDA~\cite{thoppilan2022lamda}, have been shown to be able to conduct multi-step reasoning under the chain of thought prompting~\cite{wei2022chain} or proper simple prompting template~\cite{kojima2022large}. Although the generated `reasoning' statements potentially benefit some downstream tasks, there is currently no evidence that the statements are generated via deductive reasoning, rather than obtained via pure memorisation. Generative reasoning models are difficult to evaluate since this requires huge effort of manual assessment~\cite{bostrom2021flexible}.

Although some research has demonstrated that PLMs can learn to effectively perform inference which involves taxonomic and world knowledge, chaining, and counting~\cite{talmor2020leap}, preliminary experiments on a single test set in more recent research have revealed that fine-tuning PLMs for editing knowledge might negatively affect the previously acquired knowledge~\cite{de2021editing}. Our work performs systematic and controlled examinations of the deductive reasoning capabilities of PLMs and reaches beyond (sometimes misleading) task performance.

\section{Deductive Reasoning}

\noindent \textbf{What is Deductive Reasoning?}
Psychologists define reasoning as a process of thought that yields a conclusion from precepts, thoughts, or assertions~\cite{johnson1999deductive}. Three main schools describe what people may compute to derive this conclusion: relying on factual knowledge~\cite{anderson2014rules, newell1990unified}, formal rules~\cite{braine1998steps, braine1991theory}, mental models~\cite{johnson1983mental}, or some mixture of them~\cite{falmagne1995deductive}. Our experimental study focuses on a `computational' aspect of reasoning --- namely, whether computational PLMs for reasoning inadequately generalise learned logic rules and perform inconsistently against simple adversarial reasoning examples.

We investigate deductive reasoning in the context of NLP and neural PLMs. In particular, the goal of this deductive reasoning task is to train a PLM (e.g. BERT) over some reasoning examples (each with a set of premises and a conclusion) to become a potential reasoner (e.g. R-BERT as illustrated in Figure~\ref{fig:fig1}). Then, the trained reasoner can be used to infer deductive conclusions consistently over explicit premises, where the derived conclusions are usually unseen during the PLM pretraining/training. This inference process requires the underlying PLMs to understand factual knowledge and interpret rules intervening in the deduction process. In this paper, we only focus on the encoder-based PLMs (e.g. BERT and RoBERTa) as they can be evaluated under more controllable conditions and scrutinised via automatic evaluation. In particular, we investigate two task formulations of the deductive reasoning training: \textbf{1)} \emph{classification-based} and \textbf{2)} \emph{prompt-based reasoning}, as follows.

\subsection{Classification-based Reasoning}

The classification-based approach formulates the deductive reasoning task as a sequence classification task. Let $\mathcal{D} = \{\mathcal{D}^{(1)}, \mathcal{D}^{(2)},\cdots,\mathcal{D}^{(n)}\}$ be a reasoning dataset, where $n$ is the number of examples. Each example $\mathcal{D}^{(i)}\in\mathcal{D}$ contains a set of premises $\mathcal{P}^{(i)} = \{\mathbf{p}^{(i)}_{1},\mathbf{p}^{(i)}_{2}\ldots\mathbf{p}^{(i)}_{j}\}$, a hypothesis $\mathbf{h}^{(i)}$, and a binary label $\mathbf{l}^{(i)} \in\{0,1\}$. A classification-based reasoner takes the input of $\mathcal{P}^{(i)}$ and $\mathbf{h}^{(i)}$, then outputs a binary label $\mathbf{l}^{(i)}$ indicating the faithfulness of $\mathbf{h}^{(i)}$, given that $\mathcal{P}^{(i)}$ is hypothetically factual.

\begin{figure}[t]
    \centering
    \includegraphics[width=\columnwidth]{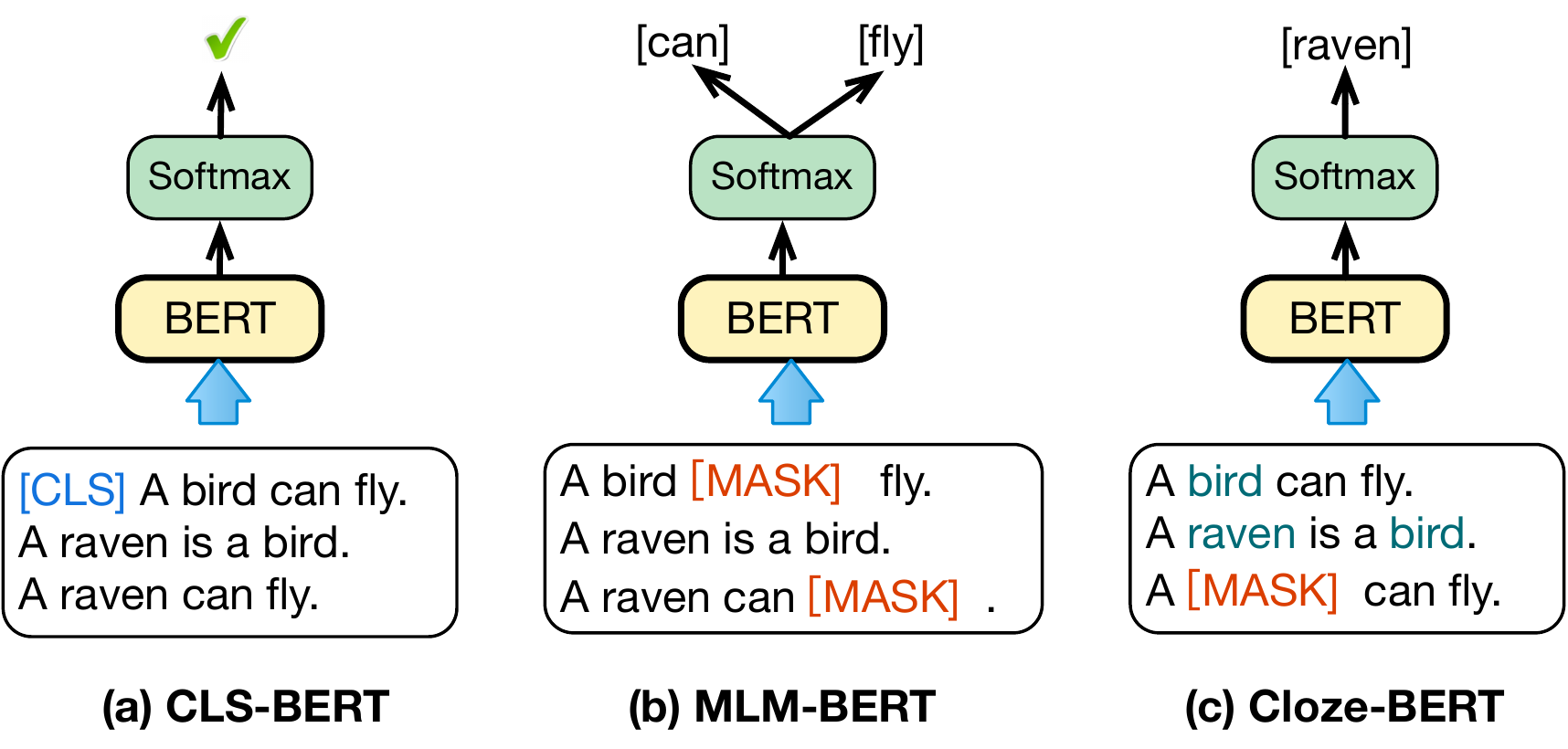}    \caption{Different reasoning training approaches.}
    \label{fig:classification}
\end{figure}

The goal of the classification-based reasoning training is to build a statistical model parameterised by ${\theta}$ to characterise $P_{\theta }( \mathbf{l}^{(i)} | \mathbf{h}^{(i)},\mathcal{P}^{(i)})$. Those PLMs built on the transformer encoder architecture, such as BERT \cite{devlin-etal-2019-bert} and RoBERTa~\cite{liu2019roberta}, can be used as the backbone of such a classification-based reasoner. Figure \ref{fig:classification}(a) shows an example of using the BERT model to train a classification-based reasoner (CLS-BERT). In particular, given a training example $\mathcal{D}^{(i)}=\{\mathbf{l}^{(i)}, \mathbf{h}^{(i)},\mathcal{P}^{(i)}\}$, the BERT model is trained to predict the hypothesis label by encoding $[\mathbf{h}^{(i)};\mathcal{P}^{(i)}]$  and computing $P_{\theta }( \mathbf{l}^{(i)} | \mathbf{h}^{(i)},\mathcal{P}^{(i)})$. To do so, the contextualised representation of the `\texttt{[CLS]}' token is subsequently projected down to two logits and passed through a softmax layer to form a Bernoulli distribution indicating that a hypothesis is \textit{true} or \textit{false}.

\subsection{Prompt-based Reasoning}

Deductive reasoning can also be approached as a cloze-completion task by formulating a valid conclusion as a cloze test. Specifically, given a reasoning example, i.e., $\mathcal{D}^{(i)}$ with its premises $\mathcal{P}^{(i)}$, and a cloze prompt $\mathbf{c}^{(i)}$ (e.g. \textit{``A \texttt{[MASK]} can fly''}), instead of predicting a binary label, this cloze-completion task is to predict the masked token $\mathbf{a}^{(i)}$ (e.g. \textit{raven}) to the cloze question $\mathbf{c}^{(i)}$.

The BERT-based models have been widely used in the prompt-based reasoning tasks~\cite{helwe2021reasoning,liu2022generated}, by concatenating the premises and the prompt as input and predicting the masked token based on the bidirectional context. In general, there are two training objectives for the prompt-based reasoning task, i.e., the mask language modelling (MLM) and task-specific (cloze-filling) objectives. For MLM, the given PLMs are trained over the reasoning examples using their original pretraining MLM objective to impose deductive reasoning ability; see Figure \ref{fig:classification}(b) for an example of the BERT reasoner MLM-BERT. For the cloze-filling objective, the PLMs are trained with a task-specific cloze filling objective. As illustrated in Figure \ref{fig:classification}(c), Cloze-BERT is trained to predict the masked token in the cloze prompt, by computing the probability $P_{\theta }( \mathbf{a}^{(i)} | \mathbf{c}^{(i)},\mathcal{P}^{(i)})$. We note that, unlike the original pretraining MLM objective where 15\% tokens of the input are masked randomly, the cloze-filling objective only masks the answer token $\mathbf{a}^{(i)}$ in the cloze prompt $\mathbf{c}^{(i)}$.

This prompt-based reasoning task matches the mask-filling nature of BERT. In this way, we can probe the native reasoning ability of BERT without any further fine-tuning and evaluate the contribution of reasoning training to the PLMs' reasoning ability. Foreshadowing, our experimental results in Section~\ref{sec:result} indicate that reasoning training impacts the model both positively and negatively.

\section{Experiments and Results}

Recent PLMs have shown surprisingly near-perfect performance in deductive reasoning~\cite{zhou2020progress}. However, we argue that high performance does not mean PLMs have mastered reasoning skills. To validate this, we run controlled experiments to examine whether PLM-based reasoners genuinely understand the natural language context, produce conclusions robustly against lexical and syntactic variance in surface forms, and apply learned rules to unseen cases.
\definecolor{myoj}{HTML}{FCCB00}
\definecolor{myblue}{HTML}{1273DE}
\definecolor{myred}{HTML}{DB3E00}
\definecolor{mygreen}{HTML}{008b02}

\begin{table*}[!t]
{\small
\begin{tabularx}{\textwidth}{X lll}

\toprule
  \textbf{Perturbation}          & \textbf{Premises  }                                              & \textbf{Conclusion}          & \textbf{Valid}\\ 
  \cmidrule(lr){2-4}
Original             & A bird can fly. A raven is a bird.                      & A raven can fly.    & \cmark\\   \cmidrule(lr){2-4}
Paraphrasing         & A bird \textcolor{myblue}{is able to} fly. A raven is a bird \textcolor{myblue}{species}. & A raven can fly.    & \cmark\\   \cmidrule(lr){2-4}
Synonym Substitution & A \textcolor{mygreen}{fowl} can fly. A raven is a \textcolor{mygreen}{fowl}.                      & A raven can fly.    & \cmark\\  \cmidrule(lr){2-4}
Negation             & A bird can fly. A raven is a bird.                      & A raven can\textcolor{myred}{not} fly. & \xmark\\   \cmidrule(lr){2-4}
Retained knowledge    & {A bird can live up to 100 years.} & {--} & \cmark \\
\bottomrule
\end{tabularx}
}%
\caption{Examples of different perturbations strategies that were used to create the adversarial dataset (see \S\ref{ss:adversarial}).}
\label{tab:examples}
\end{table*}

\subsection{Datasets}
Two datasets are used to examine the PLM-based reasoners, namely, the Leap of Thought (\textbf{LoT}) dataset~\cite{talmor2020leap} and the WikiData (\textbf{WD}) dataset~\cite{vrandevcic2014wikidata}.

\textbf{LoT} was originally proposed for conducting the classification-based reasoning experiments for deductive reasoning~\cite{talmor2020leap} and has been used as a standard (and sole) benchmark to probe the deductive reasoning capabilities of PLMs \cite{tafjord2021proofwriter,helwe2021reasoning}. This dataset is automatically generated by prompting knowledge graphs, including ConceptNet~\cite{speer2017conceptnet}, WordNet~\cite{miller1998wordnet} and WikiData~\cite{vrandevcic2014wikidata}. LoT contains 30,906 training instances and 1,289 instances for each validation and testing set. Each data point in LoT also contains a set of distractors that are similar but irrelevant to deriving the conclusion.

For the prompt-based reasoning task, we can reformulate the LoT dataset to fit our cloze-completion task. Instead of having a set of premises $\mathcal{P}$, a hypothesis $\mathbf{h}$, and a binary label $\mathbf{l}$, we rewrite the hypothesis in LoT into a cloze $\mathbf{c}$ and the answer $\mathbf{a}$ (e.g. \textit{A raven can fly.} $\rightarrow$ \textit{A [MASK] can fly.}). Note that we only generate those cloze questions on the positive examples. Consequently, the results across these two tasks are not directly comparable.

The \textbf{WD} dataset is an \textit{auxiliary} reasoning dataset which we generated and extracted from Wikidata5m~\cite{wang2019kepler}. Similar to previous work~\cite{petroni2019language, talmor2020leap}, we converted a set of knowledge graph triples into linguistic statements using manually designed prompts. The full description of the dataset construction is provided in Appendix~\ref{sec:wd_construction}. The final WD dataset contains 4,124 training instances, 413 validation instances, and 314 test instances. WD only contains positive examples: therefore, we only use this dataset for the cloze-completion task.

\subsection{Adversarial Probing}
\label{ss:adversarial}

Previous work demonstrates that PLMs can achieve near-perfect empirically results in reasoning tasks. For example, RoBERTa-based models record a near-perfect accuracy of 99.7\% in the deductive reasoning task on LoT ~\cite{talmor2020leap}. However, another recent study shows that in some natural language inference benchmarks, PLMs are still not robust to the negation examples~\cite{hossain-etal-2020-analysis}, while humans can handle negations with ease. In order to systematically probe PLMs' deductive reasoning capabilities, we design controlled experiments over three different adversarial test settings by generating different \textit{surface form perturbations}, \textit{negation of the original examples}, and creating additional \textit{retained knowledge anchors}. Table~\ref{tab:examples} shows some different adversarial examples.

\rparagraph{Surface Form Perturbations}
The theory of mental models postulates that deductive reasoning is based on manipulations of mental models representing situations. In other words, envisaging the situations and making a deduction can be viewed as a semantic process~\cite{johnson1999deductive, polk1995deduction}. On the other hand, previous works demonstrate that, instead of learning interpretable meaning representations and generalising across different surface forms, PLMs tend to learn artefacts in the training data, e.g., higher-order word co-occurrence statistics~\cite{sinha-etal-2021-masked}.

As both LoT and WD datasets are prompted from knowledge graphs, the lexical and syntactical variance of the dataset is minimal, with imaginable artefacts. To examine if the PLM-based reasoner could consistently perform reasoning against linguistic diversity and variability (in terms of both the token-level and the syntactic-level diversity), we employ two types of surface form perturbations to the data items from the original datasets:

\begin{itemize}[leftmargin=*]
   \item \textbf{Synonym Substitution:} { }In order to investigate to what extent the PLM-based reasoners would be sensitive to the token-level semantic diversity in terms of deriving their conclusions, we employ synonym substitution~\cite{dhole2021nlaugmenter} to the premises $\mathcal{P}$. Synonym substitution does not modify the syntactic structures and the premises' semantics, preserving all the original input's structural information. In our setting, a word is replaced by a uniform-randomly selected synonym based on WordNet~\cite{miller1998wordnet} with a probability of 50\%.

   \item \textbf{Paraphrasing:} { }  To further investigate the PLM-based reasoners' robustness on sentence-level semantic variability, we paraphrase the premises $\mathcal{P}$ with two paraphrasing systems: \begin{enumerate*}[label=(\roman*)]
  \item PEGASUS, an end-to-end model fine-tuned for paraphrasing~\cite{zhang2019pegasus}
  \item Syntactically Diverse Paraphrasing (SD-Paraphrasing), a two-step framework that incorporates neural syntactic preordering for better diversity~\cite{goyal-durrett-2020-neural}
\end{enumerate*}. 
\end{itemize}

\sparagraph{Negated Examples}
Understanding negation is often considered as the first test case in natural language understanding tasks~\cite{ettinger2020bert,Khemlani:2012negation,Schon:2021ki}. %
To examine whether PLMs can handle negation in the deductive reasoning task, we construct a set of negated samples by negating the hypothesis $\mathbf{h}$ or the cloze prompt $\mathbf{c}$ while keeping the premises $\mathcal{P}$ unchanged~\cite{hosseini2021understanding}. For the classification-based reasoning task, the label of the negated hypothesis is then also flipped. For the cloze-completion task, since the answer for the original query will unlikely be the same answer for the negated queries, predicting the original answer would be regarded as a wrong prediction.

\rparagraph{Anchors of Retained Knowledge} 
Prior work has shown that PLMs are prone to forgetting previously learnt knowledge when fine-tuning with new knowledge data~\cite{de2021editing}, the so-called catastrophic forgetting issue~\cite{kirkpatrick2017overcoming,Autume:2019neurips}. It is thus natural to also probe whether deductive reasoning training still retains the knowledge already stored in the original PLM. In this work, in order to measure to which extent PLMs retain the knowledge acquired during pretraining, we introduce a set of `retained knowledge statements' or \textit{anchors} for each dataset. There are two criteria for such \textit{anchors}: 
\begin{enumerate*}[label=(\roman*)]
\item they are semantically close to the conclusions in our test data, 
\item they do not meet the conditions of the premises for a reasoning replacement.\footnote{Taking the premise (a) of Figure~\ref{fig:fig1} as an example, the `bird' token in the statement `A bird can fly.' was replaced by its hyponym `raven' after reasoning training. However, in the anchor statement `The bird species is decreasing.', the `bird' token should not be replaced by `raven'.}
\end{enumerate*}
Ideally, the reasoning training should not affect the prediction of PLMs when reasoning over these anchors.

We create such a set of anchors for both LoT and WD, and investigate the behaviour of the reasoning models over these anchors based on the prompt-based reasoning task. In particular, these anchors should be real-world textual statements that contain the target word (to meet criterion (i) above), but their newly composed sentences (by the reasoning replacement) are unlikely true statements (to meet criterion (ii) above). To this end, we use the BM25 algorithm \cite{bm25} to retrieve the top 10 sentences for each `reasoning' target word from the Wikipedia corpus.\footnote{\href{https://github.com/elastic/elasticsearch}{ElasticSearch}: https://github.com/elastic/elasticsearch.} Then, we construct the anchor sentences based on these top 10 retrieved sentences by replacing their target words with their hyponym/hypernym words. The final anchors are selected from these top 10 sentences only when the newly created sentences do not likely exist in the entire Wikipedia corpus (i.e. their top-1 similar sentence should have less than a BM25 score of 50). Ideally, this set of retrained knowledge statements is relevant but should not be affected by the reasoning training.

\subsection{Evaluation}

Following previous work~\cite{talmor2020leap}, the evaluation metric for classification-based reasoning is accuracy. For prompt-based reasoning, we calculate top K recall (R@K) by measuring what fraction of the correct answers are retrieved in the top K predictions. For the negation examples, we report the top K error rate (E@K) because a retrieved answer $\mathbf{a}$ and the negated cloze question $\neg\mathbf{c}$ would compose a fallacy.

In the following, we report our findings and numerical results based on the BERT-based reasoners (in particular \texttt{bert-base-uncased}), but we note that other PLMs (such as RoBERTa) of various sizes observe the same performance trends and result in the same findings and conclusions. Appendix~\ref{sec:app_other_results} provides results for other PLMs.

\section{Results and Discussion}

\label{sec:result}

We evaluate the impact of reasoning training on the PLMs and investigate their robustness against three well-known issues of PLMs: utilising artefacts from data, incapability of modelling negation, and catastrophic forgetting. We further conduct qualitative analysis to understand the inference errors introduced by deductive reasoning training.

\begin{table}[!t]
\def\arraystretch{0.999}
\centering
{\small
\begin{tabularx}{\columnwidth}{l YYY}
\toprule
{\bf Model} & {R@1} & {R@5} & {R@10} \\
\midrule
\rowcolor{Gray}
\multicolumn{4}{c}{Dataset: \textbf{LoT}} \\
\midrule
{Pretrained} & {13.15} & {59.18} & {70.96} \\
{MLM-BERT} & {98.36} & {98.36} & {98.36} \\
{Cloze-BERT} & {99.73} & {100} & {100} \\
\midrule
\rowcolor{Gray}
\multicolumn{4}{c}{Dataset: \textbf{WD}} \\
\midrule
{Pretrained} & {27.07} & {64.97} & {72.93} \\
{MLM-BERT} & {99.04} & {99.36} & {99.36} \\
{Cloze-BERT} & {100} & {100} & {100} \\
\bottomrule
\end{tabularx}
}%

\caption{Recall of the correct answer in the top K predictions (R@K) from the BERT model before and after deductive prompt-based reasoning fine-tuning. Both MLM-BERT and Cloze-BERT achieve (near-)perfect R@1 scores after fine-tuning.}
\label{tab:1k-panlex}
\end{table}

\begin{table*}[!t]
\begin{center}
\resizebox{\textwidth}{!}{
\begin{tabular}{llllllll}
\toprule
\multirow{2}{*}{\textbf{Adversarial Probing}} & \multicolumn{3}{c}{\textbf{LoT}}                                                 &  & \multicolumn{3}{c}{WD}                                                            \\ \cmidrule(lr){2-4} \cmidrule(lr){6-8} 
                                              & Pretrained               & MLM-BERT                  & Cloze-BERT                &  & Pretrained                & MLM-BERT                  & Cloze-BERT                \\ \cmidrule(lr){2-4} \cmidrule(lr){6-8} 
Original                                      & 13.15                    & 98.36                     & 99.73                     &  & 27.07                     & 99.04                     & 100.00                    \\ \cmidrule(lr){2-4} \cmidrule(lr){6-8} 
+ Pegasus-Paraphrasing                        & 11.79 ($\downarrow$1.36) & 64.66 ($\downarrow$33.70) & 50.96 ($\downarrow$48.77) &  & 16.88 ($\downarrow$10.19) & 49.36 ($\downarrow$49.68) & 51.27 ($\downarrow$48.73) \\
+ SD-Paraphrasing                             & 5.75 ($\downarrow$7.40)  & 9.32 ($\downarrow$89.04)  & 0.55 ($\downarrow$99.18)  &  & 16.56 ($\downarrow$10.51) & 27.39 ($\downarrow$71.65) & 31.21 ($\downarrow$68.79) \\
+ Syn. Substitution                           & 20.00 ($\uparrow$6.85)   & 64.38 ($\downarrow$33.98) & 64.66 ($\downarrow$35.07) &  & 15.29 ($\downarrow$11.78) & 61.78 ($\downarrow$37.26) & 62.42 ($\downarrow$37.58) \\ \bottomrule
\end{tabular}
 }
\end{center}
\caption{R@1 scores on test sets obtained via applying various surface form perturbations from Section~\ref{ss:adversarial}.}
\label{table:find2}
\end{table*}

\subsection{Deductive Reasoning Training}

\begin{finding}
All the deductive reasoning training approaches significantly improve PLMs' reasoning capabilities, achieving near-perfect deductive reasoning performance on both the reasoning test sets.
\end{finding}

\noindent Table~\ref{tab:1k-panlex} reveals that the prompt-based reasoners achieve near-perfect performance on both datasets, regardless of the reasoning training method. In particular, on LoT dataset the R@1 score of BERT has increased from 13.15\% to 98.36\% and 99.73\% after the MLM reasoning training (i.e. MLM-BERT) and the cloze reasoning training (i.e. Cloze-BERT) respectively, which are in line with previously reported result~\cite{talmor2020leap}. The near-perfect trends are observed in the classification-based reasoning models, where CLS-BERT also achieves a high accuracy score of 94.72\% after reasoning training (Table~\ref{tab:cls-bert}).\footnote{Note that we cannot obtain the performance of original BERT since the classification head has not been trained yet.} In sum, while the off-the-shelf BERT model already demonstrates a decent level of empirical performance, conducting reasoning training on the pretrained BERT achieves strong or even near-perfect performance on both datasets.

\subsection{Surface Form Perturbations}
\definecolor{myoj}{HTML}{FCCB00}
\definecolor{myblue}{HTML}{1273DE}
\definecolor{myred}{HTML}{DB3E00}
\definecolor{mygreen}{HTML}{008b02}

\begin{table*}[!t]
\begin{center}
\resizebox{0.999\textwidth}{!}{
\begin{tabular}{llll}
\toprule
\textbf{Examples}                                                                                                                                                                                                                                                                                                                     & \textbf{BERT}      & \textbf{MLM-BERT}          & \textbf{Cloze-BERT}          \\ \hline

\begin{tabular}[c]{@{}l@{}}A \textcolor{myred}{holly} is not \textcolor{mygreen}{music}. A holly is part of a forest. \\ A \textcolor{myoj}{plant} is an \textcolor{myblue}{actor}. A \textcolor{mygreen}{music} is not an \textcolor{myblue}{actor}. \\ A \textcolor{myred}{holly} is a \textcolor{myoj}{plant}. A bluebottle is an organism. \\ A marigold is not an angiosperm. \\ A holly is not an important food source. \\ \cdashline{1-1} A \textless{}MASK\textgreater{} is not an \textcolor{myblue}{actor}.\end{tabular} & \textcolor{mygreen}{musician} (0.1213) \cmark &
\textcolor{myred}{holly} (0.7168) \xmark & \textcolor{myred}{holly} (0.9999) \xmark \\ \hline

\begin{tabular}[c]{@{}l@{}}A \textcolor{myred}{perry} is not a \textcolor{mygreen}{tree}. A \textcolor{mygreen}{tree} is not capable of burn. \\ \textcolor{myoj}{Alcohol} is capable of burn. A \textcolor{myred}{perry} is an \textcolor{myoj}{alcohol}. \\ \cdashline{1-1} A \textless{}MASK\textgreater{} is not capable of burn.\end{tabular} & \textcolor{mygreen}{tree} (0.5244) \cmark & person (0.0300) \xmark & \textcolor{myred}{perry} (0.9999) \xmark \\
\bottomrule             
\end{tabular}
}
\caption{Examples of top 1 predictions from the set of negated examples based on LoT test set. The wrong predictions are in red, and the reasonable predictions are in green. Some other key-related entities are in the same colour. The numbers in the parentheses are the respective probabilities of each prediction after the softmax layer.}
\label{neg_example}
\end{center}
\end{table*}

\begin{table}[htp]
\def\arraystretch{0.999}
\begin{center}
\resizebox{.4\textwidth}{!}{
\begin{tabular}{ll}
\toprule
\multirow{2}{*}{\bf Adversarial Probing} & \multicolumn{1}{c}{\bf LoT}  \\ \cmidrule(lr){2-2} 
                                     & CLS-BERT                  \\ \cmidrule(lr){2-2} 
Original                             & 94.72                     \\ \cmidrule(lr){2-2} 
+ Pegasus-Paraphrasing               & 83.86 ($\downarrow$10.86) \\
+ SD-Paraphrasing                    & 71.14 ($\downarrow$23.58) \\
+ Syn. Substitution                  & 84.48 ($\downarrow$10.24) \\
+ Negation                           & 9.34 ($\downarrow$85.49)  \\ \bottomrule
\end{tabular}
}
\caption{Accuracy scores on test sets obtained via applying various surface form perturbations from Section~\ref{ss:adversarial}.}

\label{tab:cls-bert}
\end{center}
\end{table}

\begin{finding} \label{f2}
Surface form perturbations drastically decrease PLMs' reasoning performance.
\end{finding}

\noindent A natural follow-up question to ask is to what extent the aforementioned near-perfect numbers really reflect the model's reasoning abilities. We thus perform surface form perturbations to add lexical and syntactic variance to the test datasets and probe the model against such variations.

Table~\ref{table:find2} demonstrates the performances of the MLM-BERT and Cloze-BERT reasoners, as well as the vanilla pretrained BERT language model, on our controlled test sets generated by different perturbation approaches. We can observe that the scores for both reasoners decrease substantially (>30\%) even when applying simple synonym substitution, which only adds lexical variance to the prompt-generated query. In contrast, the pretrained BERT language model is less vulnerable to such an issue. This finding aligns with the hypothesis that PLMs tend to memorise word co-occurrence statistics~\cite{sinha-etal-2021-masked}.\footnote{The comparison of large models (e.g. \texttt{bert-large}) can be found in Appendix~\ref{sec:app_other_results}, which indicates the similar trends.}

We also observe from Table \ref{table:find2} that the R@1 performances of all reasoners decrease markedly when applying either the Pegasus-Paraphrasing or SD-Paraphrasing methods to the premises. In particular, the drops are especially pronounced with the SD-Paraphrasing method, which is designed exactly to generate syntactically very diverse paraphrases. On the other hand, we see from Table \ref{tab:cls-bert} that similar performance degradation trends can be observed in the classification-based reasoning task. These results illustrate that current PLMs perform inconsistently against various surface form perturbations, suggesting that future work should look into the creation of more robust reasoners that should be resilient to lexical, syntactic, and semantic variability.

\subsection{Negated Examples}

\begin{finding}
All reasoners cannot distinguish between negated and non-negated examples.
\end{finding}

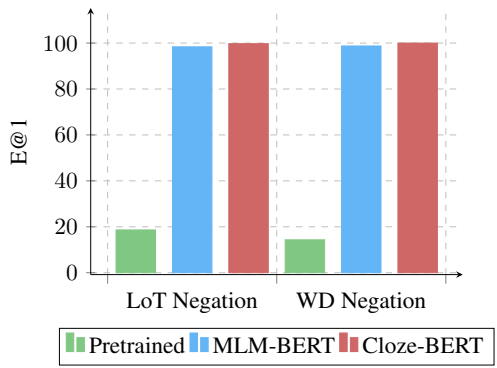
\begin{figure}[!t]
\centering

\pgfplotsset{width=\columnwidth,height=6cm,compat=1.3}

\resizebox{0.85\columnwidth}{!}{%

\begin{tikzpicture}

\begin{axis}[
	symbolic x coords={LoT Negation,WD Negation,e},
	ytick={0, 20, 40 ,60 ,80, 100},
    ybar,
    bar shift=0pt, %
	ymax=110,
	ytick={0, 20, 40 ,60 ,80, 100},
	ylabel=E@1,
	axis lines = left,
	enlargelimits=0.05,
    ymajorgrids=true,
	legend style={at={(0.5,-0.2)},
	anchor=north,legend columns=-1},
	ybar interval=0.7,
	grid style=dashed,
]

\addplot[lightgreen,fill=lightgreen] coordinates {(LoT Negation, 18.63)  (WD Negation, 14.33)  (e, 100.00) };  
\addplot[lightblue,fill=lightblue] coordinates {(LoT Negation, 98.36)  (WD Negation, 98.73)  (e, 100.00) }; 
\addplot[neuesrot,fill=neuesrot] coordinates { (LoT Negation, 99.73) (WD Negation, 100) (e, 4.46) };

\legend{Pretrained, MLM-BERT, Cloze-BERT}
\end{axis}
\end{tikzpicture}

}%
\caption{Results (E@1 scores, lower is better) on the test set comprising negated examples; before versus after deductive-reasoning training.}
\label{fig:find4}
\end{figure}

\noindent Figure~\ref{fig:find4} reveals that the error rate E@1 scores (the lower, the better)\footnote{E@1 only denotes a specific type of error: the model makes a wrong prediction due to negation being added, i.e. the negation error. However, many other types of errors that a model can make a wrong prediction, but they are not our focus and thus not measured. In other words, (1 - E@1) is not equivalent to accuracy.} for all reasoners on the set of negated examples largely increase after reasoning training. The original off-the-shelf BERT model achieves E@1 of 18.63\% on LoT and 14.33\% on the WD dataset. However, after reasoning training, the error rate of MLM-BERT significantly increases to 98.36\% (LoT) and 98.73\% (WD). Cloze-BERT's performance is even worse, with E@1 scores of 99.73\% (LoT) and 100\% (WD), suggesting a clear case of overfitting to word co-occurrence and other artefacts in the training sets. Moreover, the accuracy score on negated LoT examples is only 9.34\%, while a random baseline would score 50\%. In sum, these scores indicate that the PLM-based reasoners cannot distinguish between negated and non-negated examples, and their performance on negated examples substantially worsens after reasoning training due to task-specific overfitting.

\definecolor{myoj}{HTML}{FCCB00}
\definecolor{myblue}{HTML}{1273DE}
\definecolor{myred}{HTML}{DB3E00}
\definecolor{mygreen}{HTML}{008b02}

\begin{table*}[!t]
\begin{center}
\resizebox{.98\textwidth}{!}{
\begin{tabular}{l lll}
\toprule
\textbf{Examples   }                                                 & BERT       & MLM-BERT          & Cloze-BERT           \\ \cmidrule(lr){2-4}
a flea is a parasitic \textless{}MASK\textgreater{}.        & \textcolor{mygreen}{insect} (0.2006) \cmark   & \textcolor{myred}{substance} (0.1950) \xmark & \textcolor{myred}{drone} (0.0193) \xmark  \\ 
vapour density is a unitless \textless{}MASK\textgreater{}. & \textcolor{mygreen}{quantity} (0.3494) \cmark & \textcolor{mygreen}{quantity} (0.2197) \cmark  & \textcolor{myred}{volume} (0.0412) \xmark \\ 
both sexes have a throat \textless{}MASK\textgreater{}.     & \textcolor{mygreen}{pouch} (0.5169) \cmark    & \textcolor{myred}{\#\#lid} (0.3606) \xmark   & \textcolor{myred}{hollow} (0.0135) \xmark \\
firefox is a web \textless{}MASK\textgreater{}.             & \textcolor{mygreen}{browser} (0.9144) \cmark  & \textcolor{mygreen}{browser} (0.8861) \cmark   & \textcolor{mygreen}{browser} (0.3243) \cmark \\ 
\bottomrule
\end{tabular}
}
\caption{Examples of top 1 predictions from the set of Anchors of Retained Knowledge based on LoT. A wrong prediction is in red, and a reasonable prediction (aligning with human judgement) is in green. The numbers in the parentheses are the respective probabilities of each prediction after the softmax layer.}
\label{ret_examples}
\end{center}
\end{table*}

A quick error analysis, provided in Table \ref{neg_example}, further points to the issues with negated examples. The first example shows that the off-the-shelf BERT model makes a reasonable guess, semantically related to an entity mentioned in the premises. This guess is similar to a human guess from the same premises. However, after reasoning training, both MLM-BERT and Cloze-BERT make wrong predictions, and Cloze-BERT is extremely confident in its wrong predictions.

\subsection{Anchors of Retained Knowledge}

\begin{finding}
Previously learnt knowledge is not fully retained after reasoning training, and the trained reasoners (catastrophically) forget it.
\end{finding}

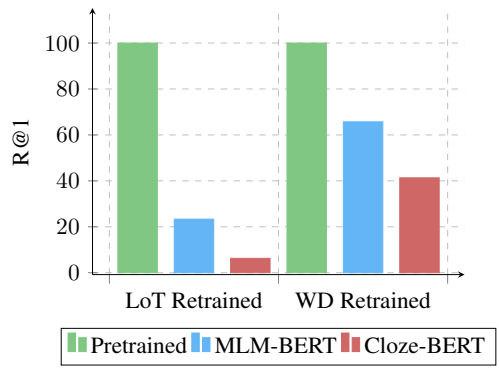
\begin{figure}[!t]
\centering

\pgfplotsset{width=\columnwidth,height=6cm,compat=1.3}

\resizebox{0.85\columnwidth}{!}{%

\begin{tikzpicture}
\begin{axis}[
	symbolic x coords= {LoT Retrained, WD Retrained, e},
	ymax=110,
	ytick={0, 20, 40 ,60 ,80, 100},
	ylabel=R@1,
	axis lines = left,
	enlargelimits=0.05,
    ymajorgrids=true,
	legend style={at={(0.5,-0.2)},
	anchor=north,legend columns=-1},
	ybar interval=0.7,
	grid style=dashed,
]

\addplot[lightgreen,fill=lightgreen] coordinates { (LoT Retrained, 100) (WD Retrained, 100) (e, 4.46) };  
\addplot[lightblue,fill=lightblue] coordinates {(LoT Retrained, 23.34) (WD Retrained, 65.7) (e, 100) };  
\addplot[neuesrot,fill=neuesrot] coordinates { (LoT Retrained, 6.27) (WD Retrained, 41.32) (e, 4.46) };

\legend{Pretrained, MLM-BERT, Cloze-BERT}
\end{axis}
\end{tikzpicture}
}

\caption{Results (R@1 scores, higher is better) on the test set comprising anchors of retained knowledge; before versus after deductive-reasoning training.}
\label{fig:find5}
\end{figure}

\noindent Figure~\ref{fig:find5} shows that performance on the anchors deteriorates substantially after reasoning training. On LoT, MLM-BERT `forgets' $\sim$77\% of the previously learnt knowledge, achieving only 23.34\% on R@1. Cloze-BERT performs even worse, scoring only 6.27\% R@1. The drops are slightly lower but still substantial on WD: MLM-BERT retains 65.7\% and Cloze-BERT retains 41.32\% of the previously stored knowledge. This result indicates that reasoning training yields the well-known phenomenon of catastrophic forgetting: this effect seems even more pronounced with Cloze-BERT, which relies on a very task-specific objective that might result in overfitting the task data.

Furthermore, Table~\ref{ret_examples} displays several qualitative examples where the predictions in green refer to the correct predictions based on human judgement. Notice that in the last example, even though the predictions from the two trained reasoners are correct, the probabilities are much lower.

Several strategies might help mitigate catastrophic forgetting. A promising direction is encapsulating lightweight adapter modules~\cite{Houlsby:2019icml,pfeiffer-etal-2020-mad, ansell-etal-2021-mad-g} within the underlying PLM, where all the `deductive reasoning capability' will be stored solely in the adapter modules, leaving the original PLM intact \cite{pfeiffer2020adapterhub}. Other similar parameter-efficient and modular methods include prefix tuning \cite{li-liang-2021-prefix} or sparse masks \cite{fisher,ansell-etal-2022-composable}. Their main premise is to separate knowledge extraction and composition, preserving previously learnt knowledge during reasoning training. We leave reasoning training with such methods for future research.

\section{Conclusion}

In this paper, we probed into the deductive reasoning capabilities of PLMs and conducted comprehensive controlled experiments to examine and compare various deductive reasoning training approaches. Our experimental results showed that current PLM-based deductive reasoners suffer from several issues: \textbf{1)} they rely on artefacts from the training data, \textbf{2)} they are incapable of modelling negation in deductive reasoning, and \textbf{3)} they forget knowledge acquired during pretraining when they get specialised into deductive reasoners. In particular, our experimental study demonstrated that models are vulnerable to multiple adversarial methods, including simple surface form perturbations such as synonym substitution or paraphrasing. While the PLMs trained for deductive reasoning achieve seemingly perfect empirical results in different reasoning datasets, they cannot yet systematically generalise to other deductive reasoning examples. Consequently, our study also calls for further, more rigorous examinations of future PLM-based models' deductive reasoning abilities. %

\section*{Acknowledgements}
Ivan Vuli\'{c} is supported by a personal Royal Society University Research Fellowship (no 221137; 2022-). Songbo Hu is supported by Cambridge International Scholarship.

\section*{Limitations}

Despite the thorough experiments on standard and popular PLMs of various sizes, this study explores only encoder-based models. Some generation-based models under other Transformer architectures, such as encoder-decoder (T5) or decoder-only (GPT-3), were also deployed in the reasoning tasks~\citep{bostrom2021flexible, wei2022chain}. We do not probe those groups of models here due to the difficult of evaluation and we leave them for future research. 

Further, for prompt-based reasoning, the current reasoners only support and have been evaluated on single-token prediction in a single language (i.e. English). Several prior works have demonstrated that conducting multi-token prediction is considerably more difficult~\citep{jiang-etal-2020-x}, which would pose an additional challenge to the PLMs in deductive reasoning tasks. One avenue of future work should also extend the scope of analyses to multilingual {and} multi-token prediction.

In addition, we note that better evaluation resources that could address paraphrases and word senses, especially for mask-filling tasks, are still lacking. This limitation is particularly significant in our setting. For example, in addition to the single-token answer in the evaluation datasets we used, there are some other feasible answers (e.g. synonyms) for the same query, which should also be considered a correct prediction. However, such answers are ignored by the current standard evaluation protocols. As a result, there is a certain level of unavoidable noise in the evaluation process.

Finally, introducing a reasoning dataset is highly challenging and appreciated by the community. Leap-of-thought is to our knowledge the only existing dataset that is suitable for our deductive reasoning evaluation. To solidify our conclusions, we further constructed an auxiliary dataset (WD) following a similar procedure to LoT. Although our data construction method is commonly used to extract reasoning examples, such an automatic procedure, unfortunately, inevitably reflects the quality and errors (e.g. nonsensical statements) from our source (WikiData). To reduce such noisy examples, we have conducted multiple rounds of filtering (see Appendix for the filtering process) and manually removed as many meaningless relations as we can, given that manually verifying each reasoning example is a highly labour-intensive task.

\bibliography{references}
\bibliographystyle{acl_natbib}

\clearpage
\appendix

\section{Experimental Details}
\noindent \textbf{Table \ref{tab:hyperparameters}} lists our model hyperparameters. Among these models, MLM-BERT and Cloze-BERT were implemented using the HuggingFace transformers package~\cite{wolf2019huggingface}. We implement CLS-BERT via the SBERT repository~\cite{reimers2019sentence}, which is built on top of the HuggingFace repository~\cite{wolf2019huggingface}. Unless mentioned otherwise, all the hyperparameters are set to the default values provided in the HuggingFace and SBERT repositories.

\begin{table}[h]
\centering
{\footnotesize
\begin{tabularx}{\linewidth}{l X}
\toprule
{\bf Hyper-parameter}                                   & {\bf Value}                         \\ \midrule
\multicolumn{2}{c}{\cellcolor[HTML]{EFEFEF}{\bf MLM-BERT and Cloze-BERT}} \\ \midrule
batch\_size                          & 2                 \\
max\_sequence\_length                              & 512                            \\
training\_epoch                          & 3                 \\

\midrule
\multicolumn{2}{c}{\cellcolor[HTML]{EFEFEF}{\bf CLS-BERT}} \\ \midrule
batch\_size                              & 64                   \\
max\_sequence\_length                     & 128                   \\
training\_epoch                          & 20                             \\\bottomrule
\end{tabularx}
}%
\caption{Model hyper-parameters.}
\label{tab:hyperparameters}
\end{table}

\section{Evaluation Results for PLMs}
\label{sec:app_other_results}

Table~\ref{table:bert_all_class_result} supplements the main paper by providing additional results with classification-based reasoners, where the reasoners start from different PLMs: \texttt{distil-bert}, \texttt{bert-base}, \texttt{bert-large}, \texttt{roberta-base}, and \texttt{roberta-large}. These results corroborate the main findings presented in the main paper; see Section~\ref{sec:result}. Larger models do perform slightly better than their smaller variants on average. However, the results also demonstrate that different models, regardless of their size, suffer from exactly the same issues, discussed in the main paper.

Table~\ref{table:appendixmlm} and Table~\ref{table:appendixcloze} demonstrate performance over adversarial test sets for prompt-based reasoners (1. MLM-based, 2. Cloze-based). The findings from these tables align with the findings from the main results presented in Section~\ref{sec:result}.

\begin{table*}[!t]
\begin{center}
\resizebox{\textwidth}{!}{
\begin{tabular}{llllll}
\hline
\multirow{2}{*}{\textbf{Adversarial Probing}} & \multicolumn{5}{c}{\textbf{LoT}}                                                                                                                   \\ \cline{2-6} 
                                     & distilbert-based-uncased  & bert-base-uncased         & bert-large-uncased        & roberta-base              & roberta-large             \\ \cline{2-6} 
Original                             & 90.22                     & 94.72                     & 98.76                     & 92.71                     & 99.38                     \\ \cline{2-6} 
+ Pegasus-Paraphrasing               & 77.27 ($\downarrow$12.95) & 83.86 ($\downarrow$10.86) & 84.40 ($\downarrow$14.35) & 80.84 ($\downarrow$11.87) & 87.20 ($\downarrow$12.18) \\
+ SD-Paraphrasing                    & 66.56 ($\downarrow$23.66) & 71.14 ($\downarrow$23.58) & 73.39 ($\downarrow$25.39) & 75.95 ($\downarrow$16.76) & 78.97 ($\downarrow$20.40) \\
+ Syn. Substitution                  & 81.46 ($\downarrow$8.76)  & 84.48 ($\downarrow$10.24) & 93.33 ($\downarrow$5.43)  & 80.76 ($\downarrow$11.95) & 94.26 ($\downarrow$5.12)  \\
+ Negation                           & 18.86 ($\downarrow$71.37) & 9.34 ($\downarrow$85.49)  & 2.58 ($\downarrow$96.28)  & 19.40 ($\downarrow$73.31) & 35.85 ($\downarrow$63.53) \\ \hline
\end{tabular}
}
\end{center}
\caption{Accuracy on different LoT adversarial test sets for classification-based reasoners with various back-boned PLMs.}
\label{table:bert_all_class_result}
\end{table*}
\begin{table*}[!t]
\begin{center}
\resizebox{\textwidth}{!}{
\begin{tabular}{llllllll}
\hline
\multirow{2}{*}{\textbf{Adversarial Probing}} & \multicolumn{3}{c}{\textbf{LoT}}                                                           &  & \multicolumn{3}{c}{\textbf{WD}}                                                            \\ \cline{2-4} \cline{6-8} 
                                     & MLM-DISTILBERT            & MLM-BERT                  & MLM-BERT-LARGE            &  & MLM-DISTILBERT            & MLM-BERT                  & MLM-BERT-LARGE            \\ \cline{2-4} \cline{6-8} 
Original                             & 99.45                     & 98.36                     & 99.45                     &  & 99.36                     & 99.04                     & 98.09                     \\ \cline{2-4} \cline{6-8} 
+ Pegasus-Paraphrasing               & 76.71 ($\downarrow$22.74) & 64.66 ($\downarrow$33.7)  & 60.27 ($\downarrow$39.18) &  & 56.05 ($\downarrow$43.31) & 49.36 ($\downarrow$49.68) & 52.23 ($\downarrow$45.86) \\
+ SD-Paraphrasing                    & 24.66 ($\downarrow$74.79) & 9.32 ($\downarrow$89.04)  & 4.93 ($\downarrow$94.52)  &  & 28.66 ($\downarrow$70.7)  & 27.39 ($\downarrow$71.65) & 33.12 ($\downarrow$64.97) \\
+ Syn. Substitution                  & 64.11 ($\downarrow$35.34) & 64.38 ($\downarrow$33.98) & 64.93 ($\downarrow$34.52) &  & 62.1 ($\downarrow$37.26)  & 61.78 ($\downarrow$37.26) & 60.19 ($\downarrow$37.9)  \\
+ Negation (E@1 $\downarrow$)        & 97.53                     & 98.36                     & 95.89                     &  & 99.36                     & 98.73                     & 98.41                     \\
Retained Knowledge                   & 25.86                     & 23.34                     & 26.67                     &  & 58.33                     & 65.7                      & 56.61                     \\ \hline
\end{tabular}
}
\end{center}
\caption{Top 1 recall (R@1) on adversarial test sets for various prompt-based reasoners with MLM training. The numbers for the negated examples indicate the top 1 error rates (the lower, the better).}
\label{table:appendixmlm}
\end{table*}
\begin{table*}[!t]
\begin{center}
\resizebox{\textwidth}{!}{
\begin{tabular}{llllllll}
\hline
\multirow{2}{*}{\textbf{Adversarial Probing}} & \multicolumn{3}{c}{\textbf{LoT}}                                                           &  & \multicolumn{3}{c}{\textbf{WD}}                                                            \\ \cline{2-4} \cline{6-8} 
                                     & Cloze-DISTILBERT          & Cloze-BERT                & Cloze-BERT-LARGE          &  & Cloze-DISTILBERT          & Cloze-BERT                & Cloze-BERT-LARGE          \\ \cline{2-4} \cline{6-8} 
Original                             & 100                       & 99.73                     & 98.63                     &  & 100                       & 100                       & 100                       \\ \cline{2-4} \cline{6-8} 
+ Pegasus-Paraphrasing               & 57.81 ($\downarrow$42.19) & 50.96 ($\downarrow$48.77) & 47.4 ($\downarrow$51.23)  &  & 57.01 ($\downarrow$42.99) & 51.27 ($\downarrow$48.73) & 56.37 ($\downarrow$43.63) \\
+ SD-Paraphrasing                    & 2.74 ($\downarrow$97.26)  & 0.55 ($\downarrow$99.18)  & 0.82 ($\downarrow$97.81)  &  & 37.26 ($\downarrow$62.74) & 31.21 ($\downarrow$68.79) & 46.82 ($\downarrow$53.18) \\
+ Syn. Substitution                  & 64.66 ($\downarrow$35.34) & 64.66 ($\downarrow$35.07) & 63.84 ($\downarrow$34.79) &  & 62.74 ($\downarrow$37.26) & 62.42 ($\downarrow$37.58) & 62.74 ($\downarrow$37.26) \\
+ Negation (E@1 $\downarrow$)        & 100                       & 99.73                     & 98.63                     &  & 100                       & 100                       & 99.68                     \\
Retained Knowledge                   & 11.79                     & 6.27                      & 1.45                      &  & 35.61                     & 41.32                     & 16.95                     \\ \hline
\end{tabular}
}
\end{center}
\caption{Top 1 recall (R@1) on adversarial test sets for various prompt-based reasoners with Cloze-filling training. The numbers for the negated examples indicate the top 1 error rates (the lower, the better).}
\label{table:appendixcloze}
\end{table*}

\section{WD Dataset Construction Pipeline}
\label{sec:wd_construction}

\begin{figure}[h]
    \centering
    \includegraphics[width=0.4\columnwidth]{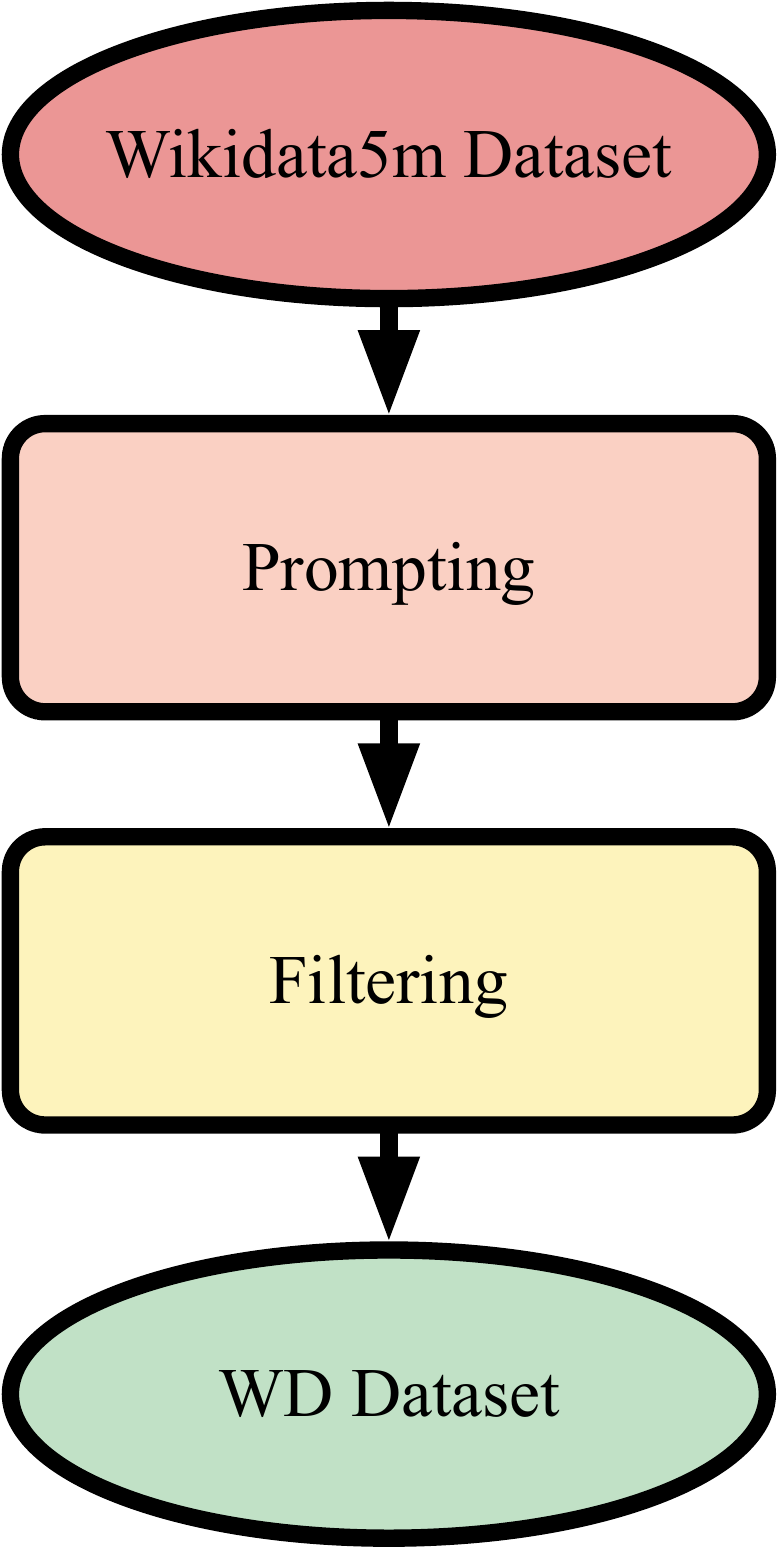}    \caption{Data construction pipeline for the \textbf{WD} dataset. This \textbf{WD} dataset is extracted and derived from WikiData5m~\cite{wang2019kepler}. Following the previous pipeline~\cite{petroni2019language}, we convert a set of knowledge graph triples into linguistic statements using manually designed prompts.}
    \label{fig:wd_pipeline}
\end{figure}

We construct the \textbf{WD} dataset following the pipeline shown in Figure~\ref{fig:wd_pipeline}, and outlined in what follows.

\rparagraph{Source Data} 
We choose the Wikidata5m dataset~\citep{wang2019kepler} as the knowledge source for \textbf{WD}. Wikidata5m is a million-scale knowledge graph dataset created upon Wikidata~\cite{vrandevcic2014wikidata}. This dataset comprises 20 million triples, describing relevant and important knowledge statements about real-world entities.

\rparagraph{Prompting} 
We manually select a set of relations based on their frequency and design their corresponding prompts shown in Table~\ref{tab:prompt}. Given a taxonomic relation, such as $is\_a$ with relation ID P31, we sample  a relevant taxonomic-knowledge graph triple: $\langle raven, is\_a, bird \rangle$. We then retrieve other relevant triples about the subject and the object, for example, $\langle bird, is\_capable\_of, fly\rangle$ and $\langle raven, is\_capable\_of, fly \rangle$. Next, we use our predefined prompts to convert the knowledge graph triples to textual knowledge statements: \textit{A bird can fly}, \textit{A bird is a raven} and \textit{A raven can fly}. Furthermore, these prompted statements assemble an inference instance in the \textbf{WD} dataset: if \textit{A bird can fly} and \textit{A bird is a raven}, then \textit{A raven can fly}. In our cloze-completion task setting, we mask the object of the taxonomic triple (e.g. $raven$) in the conclusion statement to form a cloze question (\textit{A [MASK] can fly}). Therefore, $raven$ is the correct answer to this question.

\rparagraph{Filtering}
We filter those constructed inference instances with the following properties:
\begin{enumerate*}[label=(\roman*)]
\item We only choose examples with answers being a single masked token, and these answers should be included in the BERT vocabulary.
\item For all inference instances, the maximum number of occurrences of a single answer is 50 to balance the dataset and avoid excessive repetition.
\end{enumerate*}

\rparagraph{Final WD Dataset} 
The final WD dataset contains 4,851 instances, which are randomly split into 4,124/413/314 instances for training/validation/testing while keeping that the answers of the testing set should not appear in the training/validation sets. This is to ensure that trained reasoners need to draw conclusions via conducting deductive reasoning rather than via memorisation.

\begin{table}[ht]
\begin{tabular}{@{}ll@{}}
\toprule
Relation ID & Prompt                                     \\ \midrule
\rowcolor[HTML]{EFEFEF} 
P31         & {[}X{]} is a {[}Y{]} .                     \\
\rowcolor[HTML]{EFEFEF} 
P136        & {[}X{]} is a genre of {[}Y{]} .            \\
\rowcolor[HTML]{EFEFEF} 
P179        & {[}X{]} is part of the {[}Y{]} series .    \\
\rowcolor[HTML]{EFEFEF} 
P279        & {[}X{]} is a subclass of {[}Y{]} .         \\
\rowcolor[HTML]{EFEFEF} 
P527        & {[}X{]} consists of {[}Y{]} .              \\
\rowcolor[HTML]{EFEFEF} 
P1269       & {[}X{]} is a topic of {[}Y{]} .            \\
P17         & {[}X{]} is hosted in {[}Y{]} .             \\
P39         & {[}X{]} holds a {[}Y{]} position .         \\
P101        & {[}X{]} is a subject of {[}Y{]} .          \\
P106        & {[}X{]} is a {[}Y{]} by profession .       \\
P140        & The religion of {[}X{]} is {[}Y{]} .       \\
P144        & {[}X{]} is based on {[}Y{]} .              \\
P180        & {[}X{]} is a painting of {[}Y{]} .         \\
P276        & {[}X{]} is located at {[}Y{]} .            \\
P306        & {[}X{]} runs on {[}Y{]} operating system . \\
P355        & {[}X{]} owns {[}Y{]} .                     \\
P360        & {[}X{]} is a list of {[}Y{]} .             \\
P400        & {[}Y{]} is a platform of {[}X{]} .         \\
P404        & The game mode of {[}X{]} is {[}Y{]} .      \\
P462        & The color of {[}X{]} is {[}Y{]} .          \\
P463        & {[}X{]} is a member of {[}Y{]} .           \\
P737        & {[}X{]} is influenced by {[}Y{]} .         \\
P749        & {[}Y{]} owns {[}X{]} .                     \\
P1303       & {[}X{]} plays {[}Y{]} .                    \\
P1343       & {[}X{]} is written about in {[}Y{]} .      \\ \bottomrule
\end{tabular}
\caption{Manually written prompts for generating the WD dataset. Given a triple $\langle[X], R\_ID, [Y]\rangle$, the textual knowledge statement (e.g. premises) is written based on the above prompts. R\_ID is the unique relation ID in the Wikidata5m dataset. Gray entries (first six rows) denote taxonomic relations.}
\label{tab:prompt}
\end{table}

\end{document}